\documentclass[10pt,twocolumn,letterpaper]{article}

\usepackage{cvpr}              

\usepackage{graphicx}
\usepackage{amsmath}
\usepackage{amssymb}
\usepackage{booktabs}

\usepackage{bbm}
\usepackage{color}
\usepackage{multirow}
\usepackage{subcaption}
\usepackage{pifont}

\usepackage[1]{xcolor}

\usepackage[pagebackref,breaklinks,colorlinks]{hyperref}

\usepackage[capitalize]{cleveref}
\crefname{section}{Sec.}{Secs.}
\Crefname{section}{Section}{Sections}
\Crefname{table}{Table}{Tables}
\crefname{table}{Tab.}{Tabs.}

\def\cvprPaperID{5027} 
\def\confName{CVPR}
\def\confYear{2022}

\begin{document}
	
	\title{Protecting Celebrities from DeepFake with Identity Consistency Transformer}

	\author{
		Xiaoyi Dong$^{1}$\thanks{Work done during an internship at Microsoft Research Asia.},  Jianmin Bao$^{2}$, Dongdong Chen$^{3}$\thanks{Dongdong Chen is the corresponding author.}, Ting Zhang$^{2}$, Weiming Zhang$^{1}$,\\ Nenghai Yu$^{1}$, Dong Chen$^{2}$, Fang Wen$^{2}$, Baining Guo$^{2}$ \\
		$^{1}$University of Science and Technology of China  \\
		$^{2}$Microsoft Research Asia
		$^{3}$Microsoft Cloud + AI \\
		{\tt\small\{dlight@mail., zhangwm@, ynh@\}.ustc.edu.cn } 
		{\tt\small cddlyf@gmail.com }\\
		{\tt\small\{jianbao, Ting.Zhang, doch, fangwen, bainguo \}@microsoft.com } 
	}
	
	\maketitle
	
	\begin{abstract}
		In this work we propose Identity Consistency Transformer, a novel face forgery detection method that focuses on high-level semantics, specifically identity information, and detecting a suspect face by finding identity inconsistency in inner and outer face regions. The Identity Consistency Transformer incorporates a consistency loss for identity consistency determination. We show that Identity Consistency Transformer exhibits superior generalization ability not only across different datasets but also across various types of image degradation forms found in real-world applications including deepfake videos. The Identity Consistency Transformer can be easily enhanced with additional identity information when such information is available, and for this reason it is especially well-suited for detecting face forgeries involving celebrities. \footnote{Code will be released at \url{https://github.com/LightDXY/ICT_DeepFake}}
		\vspace{-5mm}
	\end{abstract}
	
	\section{Introduction}
	
	Deepfake techniques~\cite{DeepFaceLab, DFaker,FakeApp,faceswap, faceswapGAN,li2019faceshifter,bao2018openset,nirkin2019fsgan,nirkin2018face} have been largely advanced to be able to create incredibly realistic fake images of which the face is replaced with someone else in another image. The malicious usage and spread of deepfake have raised serious societal concerns and posed an increasing threat to our trust in online media. 
	Therefore, face forgery detection is in urgent need and has gained a considerable amount of attention recently. 
	
	Notably, the overwhelming majority among all cases of face forgeries involve politicians, celebrities and corporate leaders, as their photos and videos are easier to find on the web and hence easily manipulated to generate impressively photo-realistic deepfake.
	Yet previous detection algorithms make predictions about the forgery based only on the suspect images,
	and neglect to exploit those freely available data.
	In this paper, we 
	argue that the images / videos available online can not only be used in generating face forgeries but also be utilized to detect them, and try to protect people whose face is accessible online and thus vulnerable for face manipulation, 
	i.e., celebrities in a broad sense.
	
	Recently, numerous efforts have been devoted to detecting face forgery and
	achieve promising detection performance.
	Most existing methods aim to discriminate fake images by 
	exploiting low-level textures and searching for the underlying generation artifacts \cite{zhou2018twostream,zhou2018learning,liu2018image,bappy2019hybrid,faceforensics,li2019exposing,afchar2018mesonet,matern2019exploiting,nguyen2019multitask,nguyen2019use,li2020face,qian2020thinking}. 
	While deploying these techniques in real-world products, we observe two common problems: (1) deepfake detection is usually performed on suspected videos and the video frames have image degradations, such as image rescale, noise and video codec conversion; and (2) when the generated deepfake is convincingly photo-realistic, the low-level traces of forgeries become very hard to detect. These problems make the deepfake detection unstable with video input. We wish to make deepfake detection significantly more robust by making heavy use of semantically meaningful identity information.

	\begin{figure}[t]
		\centering
		\includegraphics[width=1\columnwidth]{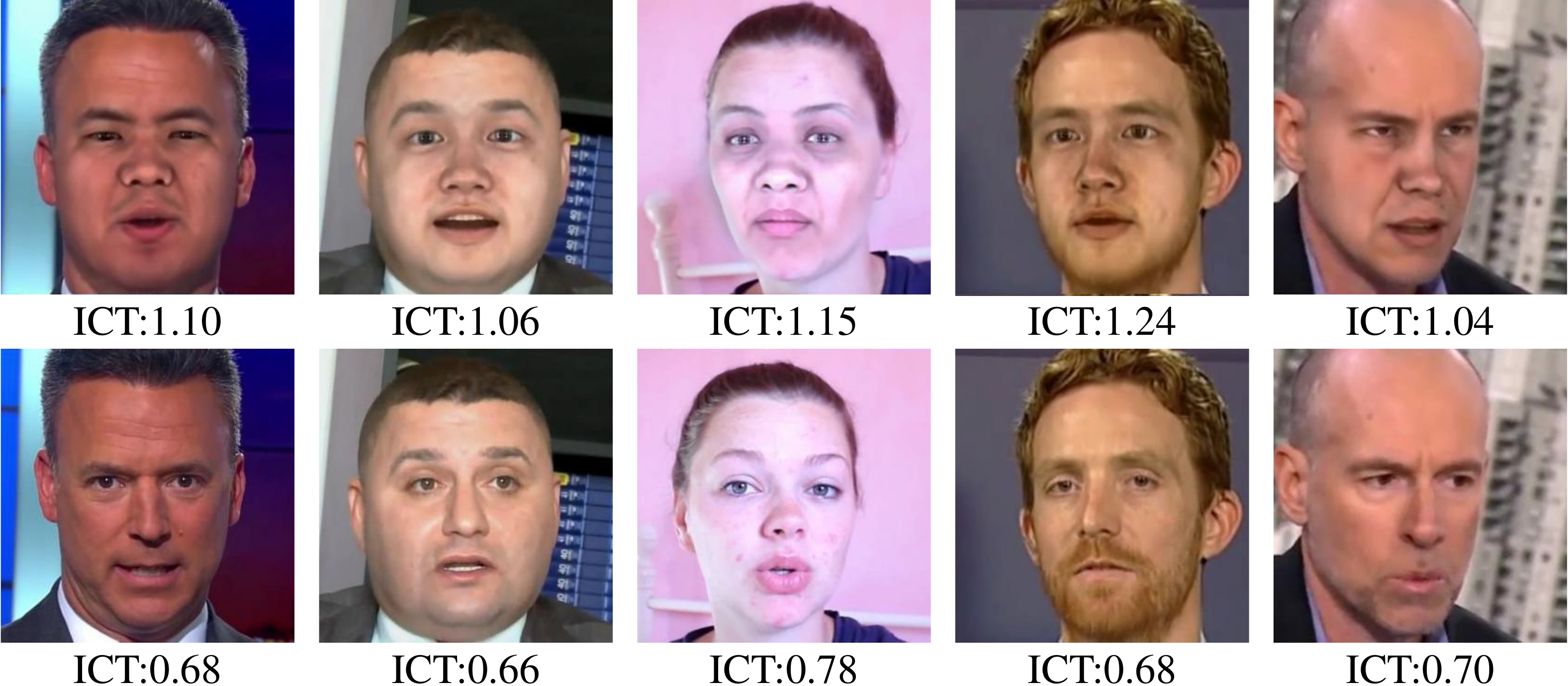} 
		\vspace{-8mm}
		\caption{Five fake (1st row) and real (2nd row) faces and their identity consistency scores. Higher ICT score indicates the inner and outer face more likely from two people, \ie DeepFake face.}
		\label{fig:intro}
		\vspace{-6mm}
	\end{figure}

	\begin{figure*}[t]
		\centering
		\includegraphics[width=2.08\columnwidth]{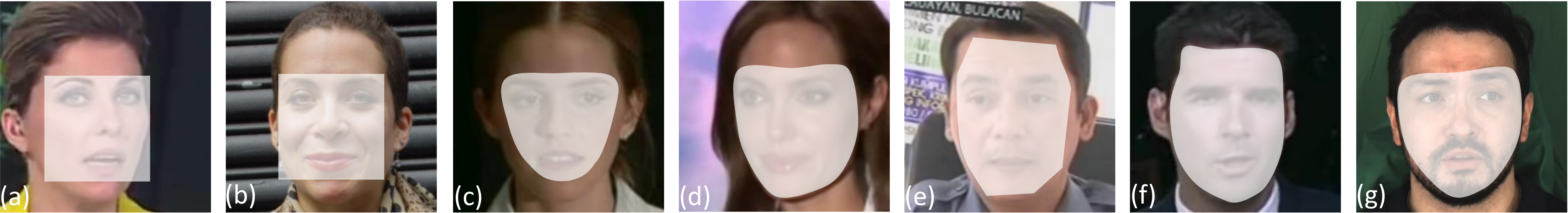} 
		\vspace{-3mm}
		\caption{Forgery regions of current face forgery methods. (a) DeepFake in FF++. (b) DeepFake in Google Deepfake Detection. (c) DeepFake in CelebDF. (d) DeepFake in DeepFaceLab. (e)Face2face. (f) FSGAN (g) DF-VAE. They replace the inner face region with different shapes while keep the outer face unchanged.}
		\label{fig:mask}
		\vspace{-6mm}
	\end{figure*}

	In this paper we propose a new face forgery detection technique called Identity Consistency Transformer (ICT) based on high-level semantics. The key idea is to detect identity consistency in the suspect image, i.e., whether the inner face and the outer face belongs to the same person. This turns out to be a non-trivial task. A naïve solution would be to compare identity vectors extracted using the off-the-shelf face recognition model from the inner- and outer- faces, similar to the popular face verification technique. Unfortunately existing face verification methods~\cite{deng2019arcface,wang2018cosface} tend to characterize the most discriminative region, i.e., the inner face for verification and fail to capture the identity information in the outer face. With Identity Consistency Transformer, we train a model to learn a pair of identity vectors, one for the inner face and the other for the outer face, by designing a Transformer such that the inner and the outer identities can be learned simultaneously in a seamlessly unified model. Our Identity Consistency Transformer incorporates a novel consistency loss to pull the two identities together when their labels are the same and consequently push them away when they are associated with different labels.

	Our approach only assumes the identity discrepancy and thus can be trained with a large number of swapped faces obtained from swapping different identities, without any fake images generated by face manipulation methods. Empirically we show that Identity Consistency Transformer exhibits significantly superior performance in the situations where low-level texture based methods fail. Figure~\ref{fig:intro} presents five examples illustrating that state-of-the-art detection methods~\cite{li2019exposing,afchar2018mesonet,faceforensics} fail to detect them while our method uncovers quite different identity consistency scores and thus can differentiate real and fake images.
	
	Another advantage of Identity Consistency Transformer is that it can be easily enhanced with additional identity information when such information happens to be available, as is the case with celebrities. For celebrities, their reference images are available and with the Identity Consistency Transformer we construct an authentic reference set consisting of the extracted identity vector pairs of the celebrities and thus create a new consistency score to enhance identity consistency detection. The resulting reference-assisted ICT (ICT-Ref) achieves the state-of-the-art performance on conventional benchmark datasets, demonstrating its strong ability for additional protection for celebrities. Finally, we show in our experiments that the proposed ICT and ICT-Ref significantly improve the generalization ability in two directions: 1) across different datasets and 2) more importantly, across different image degradation forms in real-world applications including video applications.
	
	\begin{figure*}[t]
		\centering
		\includegraphics[width=2\columnwidth]{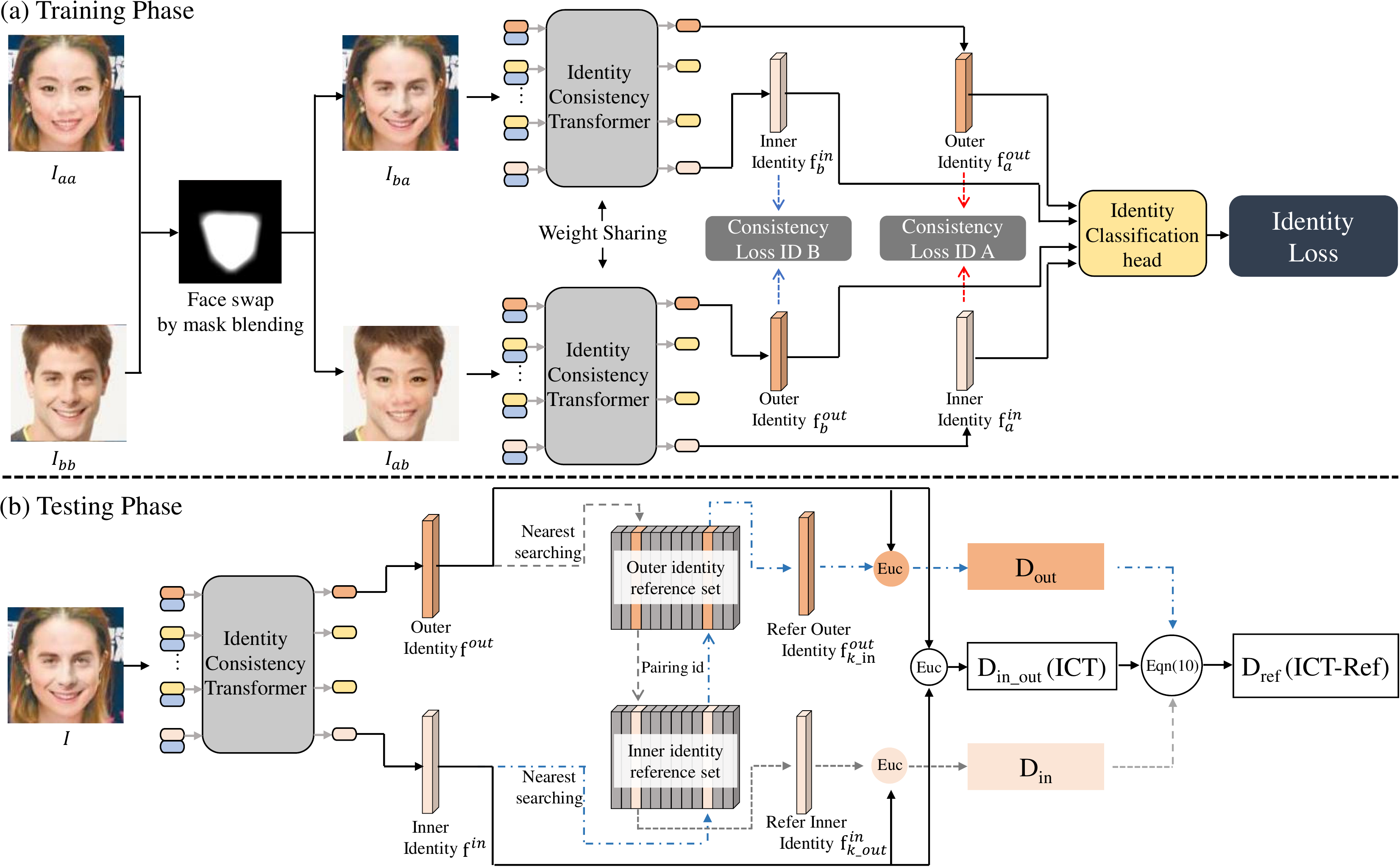}
		\vspace{-3mm}
		\caption{Illustrating the (a) training phase and (b) testing phase of our proposed Identity Consistency Transformer.}
		\label{fig:pipeline}
		\vspace{-5mm}
	\end{figure*}

	\section{Related Work}
	\noindent \textbf{Deepfake generation}.
	We roughly divide the
	face manipulation algorithms into three categories.
	(1) Early attempts are landmark-based approaches~\cite{bitouk2008face,wang2008facial} that utilize face alignment to find a source face with a similar pose to the target face and swap to the target face after adjusting the color, lighting and so on. As a consequence, these works are limited to swapping faces with similar poses.
	(2) To address that limitation, subsequent efforts~\cite{cheng20093d,dale2011video,lin2012face,nirkin2018face,nirkin2019fsgan,geng20193d,cao2013facewarehouse} 
	introduce 3D face representations, such as 
	Face2Face approach~\cite{thies2016face2face}, face reenactment~\cite{suwajanakorn2017synthesizing} and expression manipulation~\cite{masi2016we,masi2019face,clarke1996bringing}.
	However, these methods are unable to generate parts that do not exist in the source image, e.g., teeth, resulting in non-realistic artifacts in synthesized faces.
	(3) Recently GAN-based methods achieve more vivid face swapping results. Korshunava \etal~\cite{korshunova2017fast} and DeepFakes~\cite{faceswap} train a transformation model for each target person,
	while RSGAN~\cite{natsume2018rsgan}, IPGAN~\cite{bao2018openset} and FaceShifter~\cite{li2019faceshifter} introduced subject-agnostic face swapping works. 
	
	Despite the high-quality of current deepfake generation methods,
	almost all face manipulations still pay more attention to the inner face and adopt a blending post-processing step in order to generate an uncannily realistic deepfake.
	
	\noindent \textbf{Deepfake detection}.\label{sec:detection}
	With the popularity and accessibility of deepfakes, how to detect such compelling deepfakes has become increasingly critical
	and attracted a huge number of detection methods~\cite{zhao2021multi,zhou2018twostream,zhou2018learning,liu2018image,bappy2019hybrid,faceforensics,li2019exposing,afchar2018mesonet,matern2019exploiting,nguyen2019multitask,nguyen2019use,li2020face,qian2020thinking,yang2018exposing,li2018ictu,Zheng_2021_ICCV, Le_2021_ICCV,samarzija2014approach,banerjee2017srefi} recently. 
	Most of these methods are based on low-level textures to detect the visual artifacts, such as pixel-level artifacts~\cite{bappy2019hybrid,faceforensics}, texture differences~\cite{li2019exposing} and blending ghost~\cite{li2020face}.
	However, such low-level textures are easily faded away when encountering various disruptive forms of image degradation, leading to the drastic drop of the performance of those low-level based methods.
	
	In order to be more resilient, we turn to leverage more semantically meaningful traces. Along this direction, there exist some deepfake video detection methods seeking physical unnaturalness, e.g., utilizing head pose inconsistencies between the inner face and the surrounding using 3-D geometry~\cite{yang2018exposing} and detecting the abnormally low or high number of eye blinks~\cite{li2018ictu}.
	
	In this work, we are interested in exploiting the identity information, another type of high-level semantics for more robust face forgery detection.
	The most related works are~\cite{cozzolino2020id} and~\cite{nirkin2020deepfake}. 
	The first work~\cite{cozzolino2020id} proposes example-based forgery detection to detect fake videos given by a reference video,
	while our method is able to make predictions based only on the suspect image.
	Even in the reference-assisted variant, the reference is automatically retrieved, not given by a user.
	The second work~\cite{nirkin2020deepfake} also harvests the identity consistency between the inner face and the outer face.
	Yet our work is different and has three advantages:
	1) our model does not need any further training after obtaining the identity extraction network while their model does;
	2) our model does not require any fake videos generated by face manipulation methods while their model does;
	3) we introduce an extra novel and effective identity consistency loss instead of only adopting face recognition loss as in~\cite{nirkin2020deepfake}.
	Empirically, our work performs much better results, e.g., $85.71\%$ on Celeb-DF dataset, nearly  $20\%$ higher than the number in~\cite{nirkin2020deepfake}.

	\noindent \textbf{Transformer.}
	Transformer~\cite{vaswani2017attention} based entirely on a self-attention mechanism is first proposed in modeling text and has been shown enormous success in many NLP taks~\cite{devlin2018bert,radford2018improving,radford2019language,brown2020language}. Recent works~\cite{dosovitskiy2020image,chen2020generative,wan2021high,dong2021cswin,chen2021mobile,wang2021bevt} have extended Transformer to image and also achieved excellent results in vision tasks, e.g., image classification in ViT~\cite{dosovitskiy2020image,dong2021cswin,chen2021mobile}, image inpainting in \cite{wan2021high}, image generation in iGPT~\cite{chen2020generative} and video recognition \cite{wang2021bevt}. 
	
	We show that Transformer also works in face classification, a type of fine-grained image classification. 
	While it seems like a natural extension, it is actually significant. As it is generally believed that discriminating fine-grained categories depends heavily on the subtle and local differences, we demonstrate that Transformer with all global attentions can also learn semantically meaningful features for fine-grained classes.
	In addition to that, we specially modify Transformer by introducing a new consistency loss to make it appropriate for identity consistency detection.

	\section{Identity Consistency Transformer}
	Given an input face image $I$, the goal of deepfake detection is to classify the input $I$ either as a real image or a deepfake.
	We start by introducing the identity extraction model and then describe how to exploit the identity consistency to differentiate the given input.
	An overview of our approach is illustrated in Figure~\ref{fig:pipeline}.
	
	\subsection{Identity Extraction Model}
	\label{sec:model}
	
	To extract the identity information of the inner face as well as the outer face, we draw inspiration from
	the recent success of Transformer in application to full-sized images~\cite{dosovitskiy2020image, chen2020generative},
	and apply a vision Transformer~\cite{dosovitskiy2020image} on face classification to learn the identity information.
	This brings an advantage that the inner identity and the outer identity can be learned simultaneously by a uniform model instead of adopting two separate face recognition models that each is accountable for one of the two identities.
	In addition, we design a new consistency loss to make the Transformer more appropriate for identity consistency detection.

	Similar to the class token in~\cite{dosovitskiy2020image,devlin2018bert},
	we add two extra learnable embedding tokens, denoted as ``inner token'' and ``outer token'', to the sequence of patch embeddings. Each patch embedding is flattened from an image patch.
	Specifically, if $H$ and $W$ are the height and the width of the image,
	and $P$ is the resolution of the image patch, 
	the number of patch embeddings would be $N = HW/P^2$
	and the dimension of the patch embedding is $P^2C$ where $C$
	is the channel number of the image.

\begin{figure}[t]
		\centering
		\includegraphics[width=1\columnwidth]{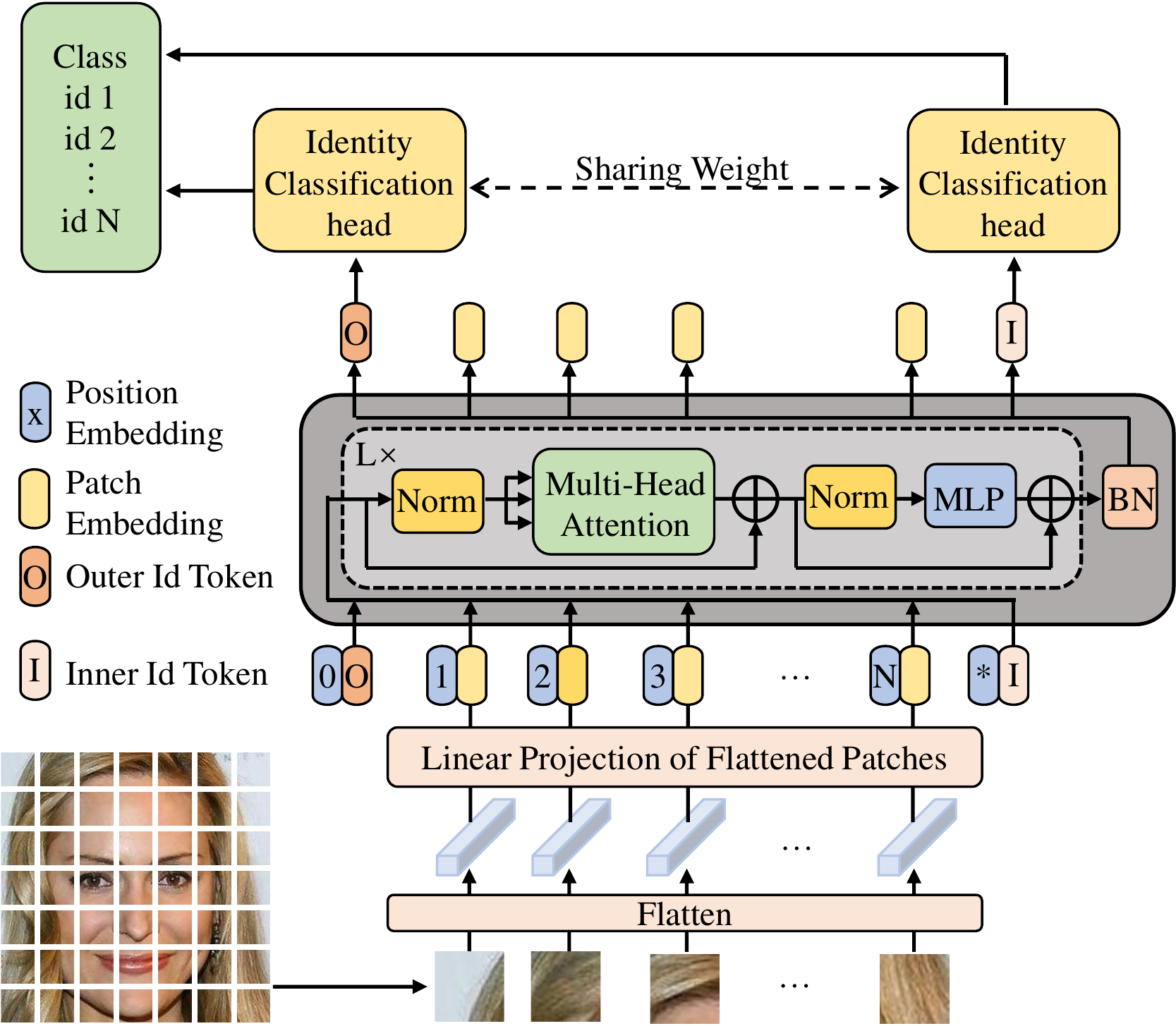}
		\vspace{-7mm}
		\caption{Architecture of the identity extraction model.}
		\label{fig:ict}
		
		\vspace{-6mm}
	\end{figure}

	Our Transformer encoder~\cite{vaswani2017attention} consists of multiple blocks that are stacked on top of each other.
	Each block includes a multi-head self-attention layer and a MLP layer with layer normalization in front.
	We illustrate the overview of model architecture in Figure~\ref{fig:ict}.
	The outputs of the inner token and the outer token of the last block are regarded as the identity information,
	which are fed into a classification head implemented by a fully connected layer.

	\noindent \textbf{Training data generation.}
	To train the identity extraction model,
	we first need real face images that are associated with identities.
	We therefore adopt the publicly available MS-Celeb-1M~\cite{guo2016ms} which is a large dataset containing 10 million face images
	and 1 million identities.
	Motivated by Face X-ray~\cite{li2020face},
	we create deepfakes by swapping the inner face of two real faces belonging to different identities, since our approach is based purely on the identity consistency to detect face forgeries. As a result, our network can be trained without fake images generated by any of the face manipulation methods.
	The generation process is similar to that in Face X-ray~\cite{li2020face}, which consists of three steps:1) find two faces that has similar poses using face landmarks;
	2) generate a mask deformed from the convex hull of the face landmarks;
	3) swap the inner face according to the mask after applying color correction.
	Except that we use local means of the RGB channels instead of the global mean of the RGB channels in the color correction technique.

	In our experiments, we only use such swapped faces during training.
	Formally, we denote the training dataset as $\mathcal{X} = \{I_{ij}\}_{i\neq j}$, where $I_{ij}$ is a fake face of which the inner face come from $I_i$ and the outer face come from $I_j$, accordingly
	$y_i$ is the inner face label and $y_j$ is the outer face label.

	\noindent \textbf{Loss functions.}
	We train the identity extraction model similar to general face classification.
	Inspired from face recognition, 
	we use the cosine-based softmax loss proposed in ArcFace \cite{deng2019arcface} rather than the vanilla cross-entropy loss,
	\begin{equation}
	\mathcal{L}_{ident} = -log \frac{e^{s \cdot cos(\theta_{i,j} + \mathbbm{1}\{j = y_i\} \cdot m)}}{\sum_{j=1}^{n}e^{s \cdot cos(\theta_{i,j} + \mathbbm{1}\{j = y_i\} \cdot m)}}.
	\end{equation}
	Here $\theta_{i,j}$ is the angle between $W_j$ and $f_i$. $W_j \in \mathbbm{R}^d$ is the $j$-th column of weights of the last fully-connected layer in the projection head, $f_i \in \mathbbm{R}^d$ is the feature of $i$-th sample in the input batch. $\mathbbm{1}\{j = y_i\}$ is the indicator function that returns 1 when $j = y_i$ and 0 otherwise. $s$ is a scale hyperparameter and $m$ is an additive angular margin. By default, we set $s=64$ and $m=0.3$.
	
	Note that the classification head is shared between the inner identity and the outer identity.
	In this way, when the inner face and the outer face belong to the same person, their corresponding identity vectors are similar
	because of the shared classification weights. On the other hand,
	the inner face and the outer face from different identities 
	are consequently pushed away.
	
	To further encourage this desired property, we introduce a consistency loss across the pair of the swapped faces, 
	\begin{align}
	\mathcal{L}_{consist} & = \mathbbm{1}\{p_{ij}^{in} = p_{ji}^{out} = y_i\}\|\mathbf{f}_{ij}^{in} - \mathbf{f}_{ji}^{out}\|_2^2 \\
	& + \mathbbm{1}\{p_{ij}^{out} = p_{ji}^{in} = y_j\}\|\mathbf{f}_{ij}^{out} - \mathbf{f}_{ji}^{in}\|_2^2.
	\end{align}
	Here $\mathbf{f}_{ij}^{in}$ ($\mathbf{f}_{ij}^{out}$) is the inner (outer) identity of image $I_{ij}$
	and
	$p_{ij}^{in}$ ($p_{ij}^{out}$) is the predicted inner (outer) label of image $I_{ij}$.
	Note that the consistency loss is activated only when their corresponding predicted labels are correct so that the consistency loss is enforced on the meaningful identity vectors.
	
	Finally, our overall loss function is,
	\begin{align}
	\mathcal{L} = \mathcal{L}_{ident} + \eta \mathcal{L}_{consist},
	\end{align}
	where $\eta$ is the weight parameter balancing the two losses.

	\subsection{Identity Consistency Detection}
	Let $T$ be the already learned identity extraction model.
	Given an image $I$,  
	we denote the extracted two identity vectors are $\mathbf{f}_{in}$ and $\mathbf{f}_{out}$, i.e.,
	$(\mathbf{f}^{in},  \mathbf{f}^{out})  = T(I)$.
	To discriminate the image, we directly compute the distance from inner to outer and use it as the measure,
	\begin{align}
	D_{in\_out} = d(\mathbf{f}^{in}, \mathbf{f}^{out}).
	\end{align}
	If $D_{in\_out}$ is smaller, this suggests that the identity is more consistent and thus the suspect image is more likely to be real.
	One may notice that as the inner identity is extracted from the inner face while the outer identity is extracted from a different region, i.e., the outer face, it is possible that $D_{in\_out}$ may still be large even if the inner face and the outer face represent the same person. 
	In fact, the proposed consistency loss that pulls the inner identity and the outer identity together when they come from the same person
	help mitigate this issue.

	\noindent \textbf{Reference-assisted identity consistency detection.}
	It is worth noting that creating deepfakes usually requires real face images to take on the role of the source image or the target image.
	We thereupon think that why not leverage publicly available real face images in detecting deepfakes.
	
	To achieve that, we construct a reference set that contains the identity vector pairs of all available real images.
	Say we have $N$ real face images, $I_1, \cdots, I_N$, and the reference set is represented as
	$\mathcal{F} = \{(\mathbf{f}^{in}_n,\mathbf{f}^{out}_n) = T(I_n)\}_{n=1}^N$.
	We then use the inner identity of the suspect image, $\mathbf{f}^{in}$, to find its nearest neighbor in the reference set,
	\begin{align}
	k^{in} = \text{argmin}_{n\in 1, \cdots,N} d(\mathbf{f}^{in}, \mathbf{f}^{in}_n),
	\alpha_{in} = d(\mathbf{f}^{in}, \mathbf{f}^{in}_{k^{in}}),
	\end{align}  
	where $d(\cdot,\cdot)$ is a pre-defined distance measure.
	This indicates that the inner face of the suspect image shares the same identity as that represented by $\mathbf{f}^{in}_{k^{in}}$ in the reference set.
	Then the remaining question is whether their corresponding outer faces belong to the same identity or not.
	Specifically, 
	we retrieve the corresponding outer identity $\mathbf{f}_{k^{in}}^{out}$ in the reference set and compute the distance to the outer identity of the suspect image,
	\begin{align}
	D_{out} = d(\mathbf{f}^{out}, \mathbf{f}^{out}_{k^{in}}).
	\end{align}
	The detection process of comparing the inner identity (i.e., nearest neighbor search using the outer identity) can be similarly derived,
	and accordingly we have  
	\begin{align}
	D_{in} &= d(\mathbf{f}^{in}, \mathbf{f}^{in}_{k^{out}}),\\
	k^{out} = \text{argmin}_{n\in 1, \cdots,N} & d(\mathbf{f}^{out}, \mathbf{f}^{out}_n),
	\alpha_{out} = d(\mathbf{f}^{out}, \mathbf{f}^{out}_{k^{out}}).
	\end{align} 
	
	The final reference-assisted identity consistency detection is evaluated by the following measure that is combined from all the distances, i.e.,
	\begin{align}
	D_{ref} = \lambda D_{in\_out} + \omega(\alpha_{in})D_{out} + \omega(\alpha_{out})D_{in}, \label{eqn:ref}
	\end{align}
	where $\omega(\cdot)$ is a weight function modulating the importance of the corresponding distance
	and $\lambda = 2.5 - \omega(\alpha_{out}) - \omega(\alpha_{in})$ is a balancing parameter.
	In our experiments, we adopt the Euclidean distance as the distance measure and $\omega(x) = \frac{1}{1+exp((x -\theta)/\tau)}$. $\theta$ is a pre-defined identity similarity threshold and $\tau$ is a scale factor. In the following experiments, we set $\theta=0.5$ and $\tau=0.5$ by default.

	\subsection{Benefits and Limitations and Impacts}
	First, our approach is general to different types of face manipulation methods as the detection model is not trained to any specific type of face manipulation.
	Second, our approach is general to different types of image degradation.
	This is because the identity information we used is a high-level evidence that is more resistant to various image degradation operations.
	Third, our method depends on the strong face recognition technique, 
	which is a very mature technology that has been widely used in real-world applications. Therefore the extracted identity information
	is quite reliable and thus improves the detection performance.
	
	On the other hand, we are aware that our framework has one main limitation:
	our approach is specifically targeted for detecting fake faces that exist identity inconsistency, namely face swap results.
	That is to say, our method may fail in detecting face reenactment results where the identity is intended to keep the same.
	Robust and general DeepFake detection in real world is a complex problem, it is hard to be solved with a single model, and we think our method is an important step that focusing on face swapping.
	
	As a face-related tasks, it is inevitable to use face-related datasets. But our method only need an one-hot label for each person and no name is required. Meanwhile, our goal is protecting the identity information from malicious usage of DeepFake.

	\section{Experiment}
	\label{sec:experimnet}
	
	\noindent\textbf{Dataset.}
	The training dataset we used is MS-Celeb-1M~\cite{guo2016ms} that contains 10 million images and 1 million identities.
	The test datasets are conventional benchmark datasets to
	demonstrate the strong generalization ability of ICT.
	(1) FaceForensics++ ~\cite{faceforensics} (FF++) is a large scale video dataset consisting of 1000 original videos and the fake videos obtained by four face manipulation methods: DeepFakes~\cite{li2020face}, Face2Face~\cite{thies2016face2face}, FaceSwap~\cite{thies2016face2face} and NeuralTextures~\cite{thies2019deferred}.
	(2) DeepFake Detection~\cite{google} (DFD) including thousands of visual deep fake videos released by Google.
	(3) Celeb-DeepFake v1~\cite{li2020celebdf} (CD1) containing 408 real videos and 795 synthesized videos with reduced visual artifacts.
	(4) Celeb-DeepFake v2~\cite{li2020celebdf} extending CD1 which contains 590 real videos and 5639 synthesized video.
	(5) DeeperForensics (Deeper) ~\cite{jiang2020deeperforensics} proposing a new deepfake generation method and generate 1000 fake videos based on FF++ ~\cite{faceforensics}.
	For our reference-assisted detection, we construct a reference set by combining the real videos of all the test datasets.
	Specifically, we randomly sample 10 images for each identity in the dataset. In the following experiments, we use frame-level AUC(\%) as the performance metric.

	\noindent\textbf{Implementation detail.} 
	The identity consistency Transformer is trained from scratch at a resolution of $112 \times 112$ without any data argumentation. 
	Our Transformer consists of 12 blocks and 12 head for the multi-head self-attention. The input image is divided to $14\times 14$ patches and we project each patch to embedding features of dimension 384.
	The number of training epochs is 30, and the batch size is 1024. The initial learning rate is set to 0.0005 and divided by 10 after 12, 15, 18 epochs.
	The loss balancing weight parameter $\eta$ is set as 4 at the first epoch, and increase by 0.5 after each epoch. To further ease the training, we set the margin $m$ as 0 at the first epoch and increase to 0.3 after 10 epochs.

	\begin{table}[t]
		\begin{center}
			\small
			\setlength{\tabcolsep}{1.1mm}{
				\begin{tabular}{l|c|c|c|c|c|c}
					\hline
					Method       &  DFD  & FF++  & Deeper& CD1   & CD2 & Avg\\
					\hline\hline
					Multi-task\cite{nguyen2019multitask}   & 65.21 & {\color{gray}72.23} & 65.32 & 72.28 & 61.06 & 65.96 \\
					MesoInc4~\cite{afchar2018mesonet}     & 59.06 & 63.41 & 51.41 & 42.26 & 53.60 & 53.95 \\
					Capsule ~\cite{nguyen2019use}     & 69.70 & {\color{gray}96.50} & 68.44 & 69.98 & 63.65 & 67.94 \\
					Xcep-c0~\cite{faceforensics}  & 89.05 & {\color{gray}99.26} & 57.76 & 48.08 & 50.37 & 61.32 \\
					Xcep-c23~\cite{faceforensics} & \textbf{95.60} & {\color{gray}98.54} & 69.85 & 74.97 & 77.82 & 79.56\\
					FWA  ~\cite{li2019exposing}        & 80.59 & 74.82 & 45.46 & 72.88 & 64.87 & 67.72\\
					DSP-FWA~\cite{li2019exposing}      & 90.99 & 81.90 & 60.00 & 78.51 & 81.41 & 78.56\\
					CNNDetect~\cite{wang2020cnn}    & 60.12 & 71.08 & 57.16 & 56.12 & 57.17 & 60.33\\
					Patch-Foren~\cite{chai2020makes}  & 49.91 & {\color{gray}73.75} & 55.35 & 59.66 & 57.16 & 55.52\\
					FFD ~\cite{dang2020detection}         & 76.61 & \textbf{92.32} & 46.64 & 74.15 & 77.80 & 73.50\\
					Face X-ray~\cite{li2020face}   & 94.14 & {\color{gray}98.44} & 72.35 & 74.76 & 75.39 & 79.16\\
					Two-Branch~\cite{masi2020two}*   & - & - & - & - & 73.41 & -\\
					PCL+I2G  ~\cite{zhao2020learning}*    & - & - & - & - & 81.80 & -\\
					Nirkin.\etal ~\cite{nirkin2020deepfake}*& - & {\color{gray}99.70} & - & - & 66.00 & -\\
					\hline
					ICT (Ours)  & 84.13 & 90.22 & \textbf{93.57} & \textbf{81.43} & \textbf{85.71} & \textbf{87.01}\\
					ICT-Ref (Ours)      & 93.17 & 98.56 & 99.25 & 96.41 & 94.43 & 96.34\\
					
					\hline
			\end{tabular}}
		\end{center}
		\vspace{-6mm}
		\caption{DeepFake detection AUC on unseen datasets. Here {\color{gray}gray} means close-set evaluation (training and testing on the same dataset). `Avg' means the average open-set AUC (the close-set results and methods with only one result is not included). `*' indicates the code is not released and we report the results from the original paper.
			The best results among all the methods except ICT-Ref are denoted in bold.}
		\vspace{-4mm}
		\label{tab:unseendataset}
	\end{table}

	\begin{figure*}[t]
		\centering
		\includegraphics[width=2\columnwidth]{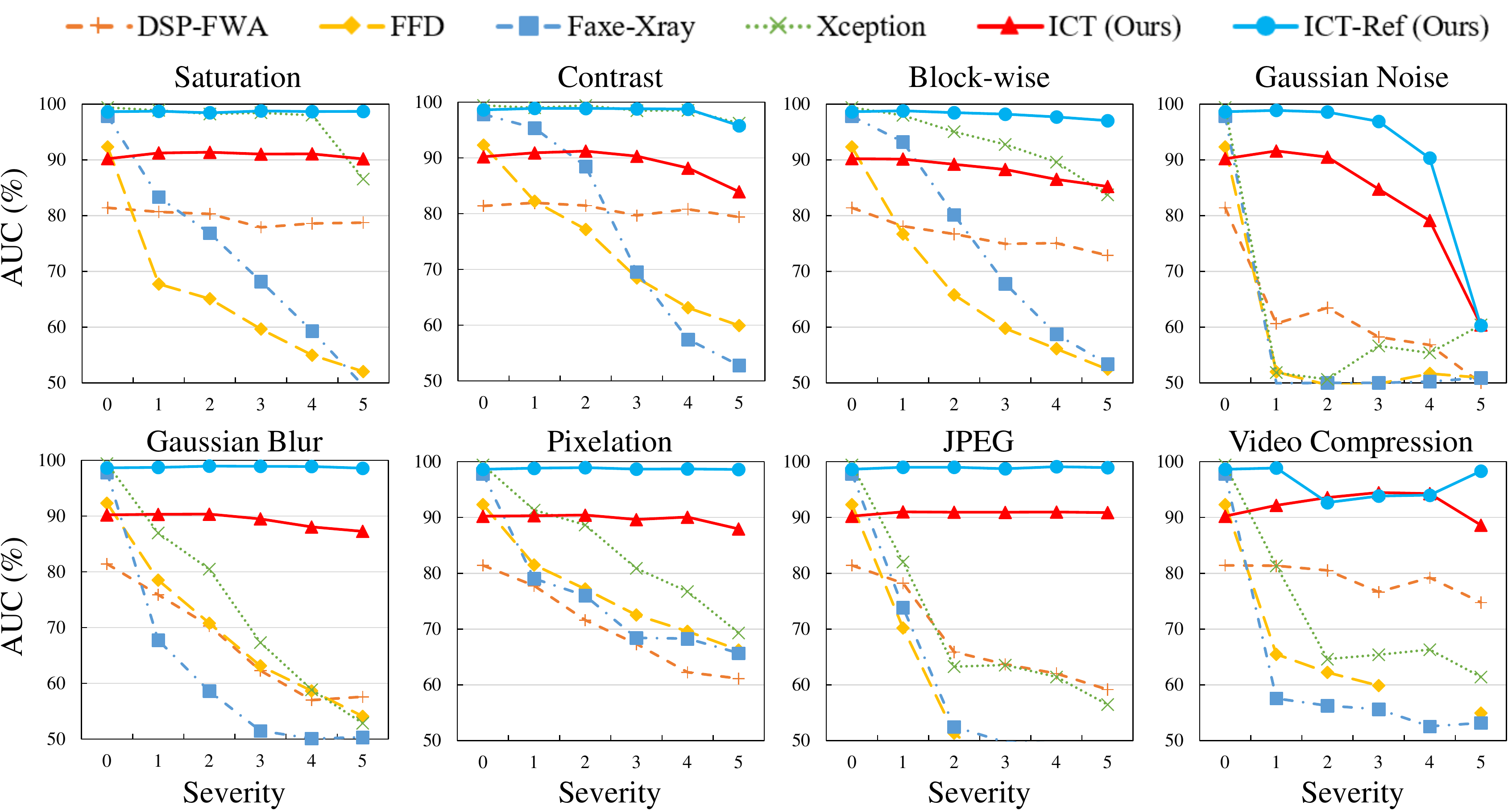}
		\vspace{-4mm}
		\caption{Generalization ability to different image degradation. Here we report frame-level AUC scores with respect to five severity level for each degradation form. }
		\label{fig:generalization}
		\vspace{-7mm}
	\end{figure*}
	
	\subsection{Comparison with State-of-the-art Methods}
	We compare our approach with state-of-the-art methods from two categories:
	methods that are specifically designed for detecting images from certain face manipulations: Multi-task~\cite{nguyen2019multitask}, MesoInc4~\cite{afchar2018mesonet}, Capsule ~\cite{nguyen2019use}, Xception-c0, c23~\cite{faceforensics}, and FWA, DSP-FWA~\cite{li2019exposing}, Two-Branch~\cite{masi2020two}, PCL+I2G  ~\cite{zhao2020learning}, Nirkin.\etal ~\cite{nirkin2020deepfake}; 
	and methods that aim to detect general deepfakes: Face X-ray~\cite{li2020face}, FFD ~\cite{dang2020detection}, CNNDetection~\cite{wang2020cnn}, and Patch-Forensics~\cite{chai2020makes}.
	Indeed, the generalization ability is the most important property of deepfake detection for real-world usage and also the biggest challenge among existing methods.
	We evaluate the generalization ability in two directions: across different unseen datasets and across different unseen image degradation forms.  
	The evaluation metric is the widely used AUC (area under the Receiver Operating Characteristic curve) and we report frame-level AUC in all experiments.

	\noindent\textbf{Generalization ability to unseen datasets.}
	In real-world scenarios, it is often the case that the suspect image is generated by an unseen face manipulation method, hence the generalization ability to unseen data is of great importance for deepfake detection. In Table~\ref{tab:unseendataset}, we present the detection performance on five different datasets, where the close-set results (\ie, training and testing are on the same dataset) is marked with {\color{gray}gray}.
	We can see that all the compared low-level based baselines perform not so good on Deeper, Celeb-DF v1 and Celeb-DF v2 datasets, which are newly released and thus exhibit less artifacts. While our approach ICT gets significant improvement and achieves high detection AUC, indicating that the identity information is a more reliable evidence than low-level textures in this case.
	On the other hand,
	our ICT-Ref variant further greatly boosts the number and achieves the state-of-the-art performance over all benchmark datasets.

	\noindent\textbf{Generalization ability to image degradation.}
	Image degradation is very common in the spread of deepfake images and videos. 
	Therefore, we further evaluate the generalization ability to different kinds of image degradation. Here we follow the image degradation strategies in ~\cite{jiang2020deeperforensics}. Note that The `JPEG' in ~\cite{jiang2020deeperforensics} is indeed Pixelation, so we realize real JPEG as an extra  degradation and set five severity level as jpeg quality factor $90, 70, 50, 30, 20$.
	
	The performance variation with respect to different degrees of all types of degradation is plotted in Figure~\ref{fig:generalization}.
	It can be easily seen that all the compared methods drop drastically when the degradation gets severer, while our approach still possesses a high level of detection performance.
	Except for the Gaussian noise where our approach also degrades when the degree gets too severe.

	\begin{table}[t]
		\begin{center}
			\small
			\setlength{\tabcolsep}{2.5mm}{
				\begin{tabular}{l|c|c|c|c}
					\hline
					
					Method &  Video1 & Video2 & Video3  & Avg\\
					\hline\hline
					Multi-task~\cite{nguyen2019multitask}   & 62.50 & 26.45 & 78.50 & 55.81 \\
					DSP-FWA ~\cite{li2019exposing}     & 23.08 & 33.14 & 51.81 & 36.01 \\
					Xception-c23~\cite{faceforensics} & 44.56 & 83.85 & 99.31 & 75.90 \\
					FFD ~\cite{dang2020detection}         & 56.04 & 71.16 & 79.07 & 68.75 \\
					Face X-ray ~\cite{li2020face}  & 66.49 & 94.03 & 87.10 & 82.54 \\
					\hline
					ICT (Ours)   & 82.27 & 97.78 & 98.05 & 94.36 \\
					ICT-Ref  (Ours)  & 100.0 & 100.0 & 100.0 & 100.0 \\
					
					\hline
			\end{tabular}}
		\end{center}
		\vspace{-6mm}
		\caption{Deepfake detection AUC (\%) on carefully crafted videos collected from Internet.}
		\vspace{-6mm}
		\label{imitate}
	\end{table}

	\noindent\textbf{Real-world scenario simulation}.
	We further showcase one typical real-world deepfake detection example by using our ICT.
	The real-word deepfake videos are downloaded from the YouTube channel ``Ctrl Shift Face\footnote{\url{https://www.youtube.com/channel/UCKpH0CKltc73e4wh0_pgL3g}}'', which are carefully crafted and all the faked identities are celebrities. 
	With the results listed in Table~\ref{imitate}, we find that the accuracy of most existing methods is relatively low. But for our ICT, we get high performance which is nearly 95\% on average. 
	The reference-assisted version even boosts the performance to 100\%. The video-version results please refer to 
	\url{https://www.youtube.com/watch?v=zgF50dcymj8}.

	\subsection{Analysis of ICT}
	
	\noindent\textbf{The effect of consistency loss.}
	In our framework, we introduce a new consistency loss to further pull the inner identity
	and the outer identity together when their corresponding labels are the same.
	Here, we experiment with an alternative counterpart, i.e., ICT without consistency loss,
	and present the comparison in Table~\ref{pose}.
	It can be seen that the performance of our model without consistency loss greatly drops by about $23\%-40\%$, which verifies that the proposed consistency loss plays a critical role in our identity consistency Transformer.

	\begin{table}[t]
		\begin{center}
			\small
			\begin{tabular}{c|c|c|c|c}
				\hline
				& DFD & FF++ &  Deeper & CD2 \\
				\hline\hline
				w/o consistency loss   & 60.82 & 62.24 & 52.90 & 51.48\\
				w/o mask deformation  & 83.59 & 83.47 & 79.06 & 83.44\\
				w/o color correction  & 80.43 & 90.11 & 92.15 & 82.86  \\ 
				ICT           & 84.13 & 90.22 & 93.57 & 85.71 \\ 
				\hline
			\end{tabular}
		\end{center}
		\vspace{-6mm}
		\caption{Analysis of consistency loss, mask deformation and color correction for our Identity Consistency Transformer.}
		\vspace{-4mm}
		\label{pose}
	\end{table}

	\noindent\textbf{The effect of different consistency measure}.
	Here we study each component in Equation~\ref{eqn:ref} and experiment four kinds of different identity consistency measures:
	(1) the proposed ICT, i.e., $D_{in\_out}$; (2) Only using $D_{in}$;
	(3) only using $D_{out}$; and (4) the proposed ICT-Ref, i.e., $D_{ref}$.
	The resulting detection AUC for the four different measures on an example dataset Celeb-DeepFake v2~\cite{li2020celebdf}  are 
	$85.71\%$, $92.52\%$, $87.45\%$ and $94.43\%$ respectively.
	As expected, the reference-assisted ICT achieves the best performance by combining the other three types.
	It is worth noting that even only using $D_{in}$ or $D_{out}$ achieves better performance than ICT.
	This shows that we have effectively made good use of the freely available reference set and verified its crucial role in detecting deepfakes.
	In addition, it is surprisingly found that using $D_{in}$ performs much better than using $D_{out}$, achieving about $5\%$ gain.
	The reason might be that as the outer face is usually not manipulated in face swapping results, 
	using the outer identity to retrieve the nearest neighbor (followed by comparing the inner identity) is more likely to retrieve the accurate suspect identity.

	\begin{figure}[t]
		\centering
		\includegraphics[width=1\columnwidth]{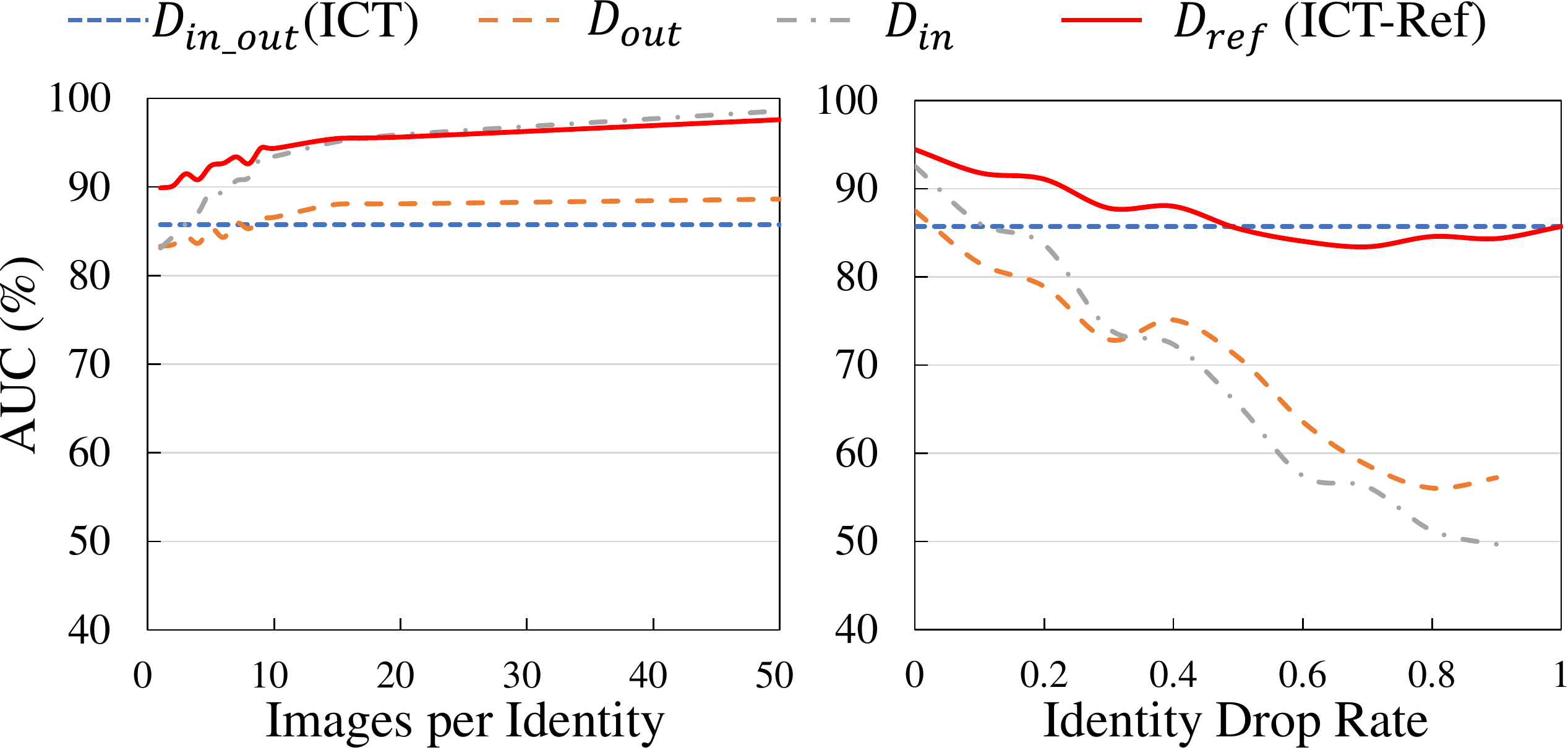} 
		\vspace{-8mm}
		\caption{Illustrating the effect of the size of the reference set by considering varying the number of images for each identity (left figure) and varying the number of identities (right figure).}
		\label{fig:pose}
		\vspace{-6mm}
	\end{figure}

	\noindent\textbf{The effect of the reference set}.
	We further analyze how the size of the reference set affects the detection performance.
	We consider two factors that affect the size of the reference set:
	(1) the number of images randomly sampled for each identity;
	and (2) the number of identities.
	The results on Celeb-DeepFake v2~\cite{li2020celebdf}  are presented in Figure~\ref{fig:pose}.
	It can be seen that in general as the number of images per identity increases, the performance also improves until get saturated,
	and as the number of identities decreases, the performance also decline.
	Interestingly, we find that at some point, e.g., small number of images per identity or larger identity drop rate, the performance gets even lower than the ICT which does not rely on the reference set.
	The reason might be that in this case, the size of the reference set is so small that the retrieved identity is not accurate.
	Note that larger identity drop rate has a stronger influence on the performance, even making the ICT-Ref lower than ICT.

	\begin{table}[t]
		\centering
		\begin{center}
			\resizebox{\linewidth}{!}{
				\setlength{\tabcolsep}{4mm}{
					\begin{tabular}{l|c|c|c}
						
						\hline
						Model&\#Params
						&  Deeper & CD2 \\
						\hline\hline
						Res50       & 43.79M & Nan & Nan \\
						Res50-split-image & 43.79M &  79.42 & 73.71\\
						Res50-split-model  & 2$\times$ 43.79M & 91.48 & 84.66 \\
						\hline
						ICT           & 21.45M &  93.57 & 85.71\\
						\hline
			\end{tabular}}}
		\end{center}
		\vspace{-7mm}
		\caption{Comparison between the proposed Transformer-based ICT and ConvNet-based Res50.}
		\label{model}
		\vspace{-5mm}
	\end{table}

	\noindent\textbf{The effect of ViT backbone.}
	Here we compare our ViT backbone with traditional CNN models ResNet50~\cite{he2016deep}. We add two separate fully connected layers following the last $7\times 7$ feature map in order to learn the inner identity and the outer identity, and finally pass the identity vector to the shared classification head (one fc similar to ours).
	
	However, we find this CNN model (denoted as Res50) fails to converge. The reason might be that the shared backbone pulls the inner identity and the outer identity so hard that it fails to capture the difference when the two identities have different labels. While for our ICT, it has the global attention mechanism at every layer so it could split the inner and outer easily. 
	We further study two substitute models: a) Res50-split-image, splitting images by the blending mask and letting the model learn inner (outer) identity from the cropped inner (outer) face rather than learning from the entire image; 
	b) Res50-split-model, using two separate models to learn the inner and outer identity individually. 
	The results are shown in Table~\ref{model}.
	We can see that Res50-split-image still performs poorly and Res50-split-model performs slightly worse than our ICT, with the cost of $4\times $ the number of parameters and two separate models.
	
	\noindent\textbf{The saliency map of the inner identity and the outer identity.}
	Here we show the attention map of the inner identity and the outer identity to see which part of the face contributes most to learning the inner identity and also the outer identity as well.
	We can see that the inner identity mainly focuses on the most discriminative inner face part, which is carefully crafted by deepfake techniques. On the contrary, the outer identity focuses on the surrounding areas, such as face contour, which is usually unchanged during the DeepFake generation. 
	We further test the LFW face recognition accuracy and find the accuray of both inner and outer token is higher than 98\%.
	With such difference between attention regions, the inner identity and the outer one are indeed semantically meaningful and thus beneficial for examining the identity inconsistency.

	\begin{figure}[t]
		\centering
		\includegraphics[width=1\columnwidth]{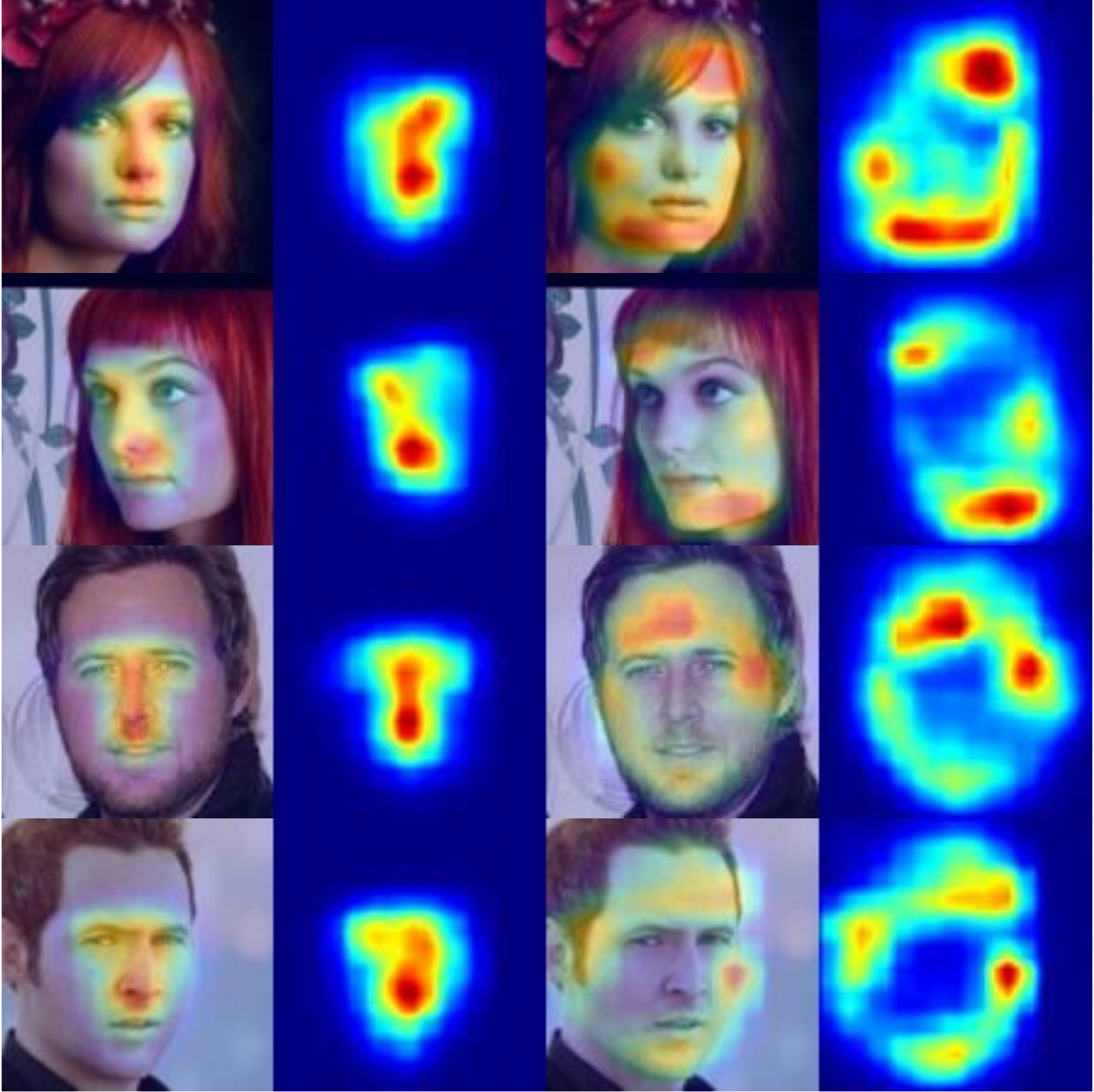} 
		\vspace{-7mm}
		\caption{Saliency map of the inner identity (the second column) and the outer identity (the fourth column) with different poses.}
		\label{fig:cam}
		\vspace{-5mm}
	\end{figure}

	\noindent\textbf{The effect of training data generation.}
	Our method is a fake-free method and only needs swapped images generated by real faces for training. 
	Here we ablate (a) mask deformation aiming to generate different shapes of mask and (b) color correction aiming to produce a more realistic swapped face in the training data generation process.
	The ablation results are given in Table~\ref{pose}.
	It can be seen that the performance of our model without mask deformation drops slightly on some of the datasets while greatly on others, as is the color correction. This verifies that both techniques are important and helpful for the final model. 
	
	\vspace{-2mm}
	\section{Conclusion}
	In this work, we propose a novel approach, identity consistency Transformer, for the detection of forged face images. 
	We adopt Transformer to simultaneously learn the inner identity as well as the outer identity and a novel consistency loss is introduced. 
	We show that our work based on high-level semantics is specially effective for the cases where low-level based methods fail.
	In addition, 
	our approach is further enhanced by leveraging additional identity information from celebrities. 
	Extensive experiments have been conducted to demonstrate the effectiveness of our approach.
	We hope that our work can encourage more works to investigate inconsistencies in high-level semantics for face forgery detection.
	
	\noindent\textbf{Acknowledgement.}
	\noindent This work was supported in part by the Natural Science Foundation of China under Grant U20B2047, 62072421, 62002334, and 62121002, Exploration Fund Project of University of Science and Technology of China under Grant YD3480002001, and by Fundamental Research Funds for the Central Universities under Grant WK2100000011.
	{\small
		\bibliographystyle{ieee_fullname}
		\bibliography{egbib}
	}
	
\end{document}